# Embedding NDRE Trajectories into Contrastive Learning for Label-Free, Physiology-Aware Crop-Stress Staging and DSS Outputs

Shafqaat Ahmad

Data Scientist, Brandt Group of Companies, Canada.

shafqaat.ahmad@brandt.ca

**Abstract**

Timely detection of crop stress is critical for sustaining yields under increasing drought frequency, yet conventional vegetation index thresholds or image-based clustering often fail to capture stress progression, limiting their value for farm decision-making. To address this gap, we present EigenCL, a physiology-guided contrastive learning framework that stages crop stress from Sentinel-2 NDRE trajectories, with the goal of providing interpretable and transferable stress diagnostics for decision support systems (DSS). EigenCL was trained on 10,000 maize NDRE patches from drought-affected Iowa fields in 2020 and tested on Nebraska fields in 2023 without retraining, with validation incorporating soil-moisture records, U.S. Drought Monitor maps, and county-level yield statistics. The model produced four physiologically coherent stress clusters (Healthy, Mild, Moderate, Severe), significantly outperforming baselines including K-Means, SimCLR, ProtoCLR, and an ablation model (Silhouette = 0.748, DBI = 0.35, CHI = 49,624). Clusters aligned with maize growth stages, with severe stress peaking around tasseling–silking (VT–R1), a stage known to drive yield loss; moreover, EigenCL clusters correlated with soil moisture at 0–14-day lags ($\rho$ up to 0.72) and matched yield anomalies in drought-affected counties. By embedding NDRE trajectory dynamics into contrastive learning, EigenCL enables early stress alerts and interpretable DSS outputs (e.g., heatmaps, scouting priorities, regional risk indices), extending beyond single-date NDRE thresholds and supporting scalable monitoring for climate-smart agronomy.

## 1. Introduction

Early detection of crop stress is critical for sustaining yields under increasing climate variability. Stress during sensitive stages, such as tasseling–silking (VT–R1) in maize, can cause yield reductions of up to 50% (Daryanto et al., 2016). For farmers, extension agencies, and insurers, timely monitoring is essential for scouting, irrigation scheduling, and risk assessment (Jones et al., 2019; Lobell et al., 2020).

Remote sensing provides scalable tools for crop monitoring, but most clustering approaches group fields by visual similarity or single-date vegetation indices (Zhang et al., 2018; Sun et al., 2022). Indices such as NDRE are physiologically meaningful because they capture canopy chlorophyll content and respond earlier than NDVI under stress (Sims et al., 2002; Fitzgerald et al., 2010; Gitelson et al., 2005; Delegido et al., 2011). However, threshold-based or static NDRE values often fail to reflect the temporal dynamics of stress progression (Peng et al., 2018). As a result, many current monitoring systems have limited decision value for farmers and policymakers.

Self-supervised learning (SSL) has recently been applied to time series without labels (Chen et al., 2020; He et al., 2020; Wang & Isola, 2020). While SSL reduces field data costs, most methods use generic similarity measures (e.g., cosine distance) or visual augmentations (Ayush et al., 2021) that can distort physiological signals. In agricultural applications, this reduces interpretability and constrains adoption (Kamilaris & Prenafeta-Boldú, 2018; Liakos et al., 2018).

To address these gaps, we introduce EigenCL (Eigenvector-Guided Contrastive Learning), a physiology-aware SSL framework that anchors similarity to the eigenvector structure of NDRE temporal trajectories. By emphasizing chlorophyll dynamics instead of static appearance, EigenCL produces clusters that align with stress-sensitive stages and can generalize across years and regions. We evaluate EigenCL using maize NDRE patches from Iowa (2020) and Nebraska (2023), validate against soil moisture, drought maps, and yield anomalies, and demonstrate how outputs can be integrated into decision support systems for climate-smart agronomy.

We evaluate EigenCL in two contexts: (1) training on Iowa 2020 NDRE patches and testing directly on Nebraska 2023 without retraining, and (2) validating clusters against soil moisture, U.S. Drought Monitor maps, and county yield records in Buchanan, Monona, and Buena Vista counties. These multi-scale checks demonstrate that EigenCL captures meaningful stress trajectories rather than noise.

Our main contributions are:

1. A physiology-aware contrastive learning framework that uses NDRE trajectories without labels.
2. A cross-year, cross-region test showing generalization from Iowa to Nebraska.
3. Multi-scale validation combining soil moisture, drought maps, and county yield data.
4. Evidence that EigenCL can support stress detection for decision support system.

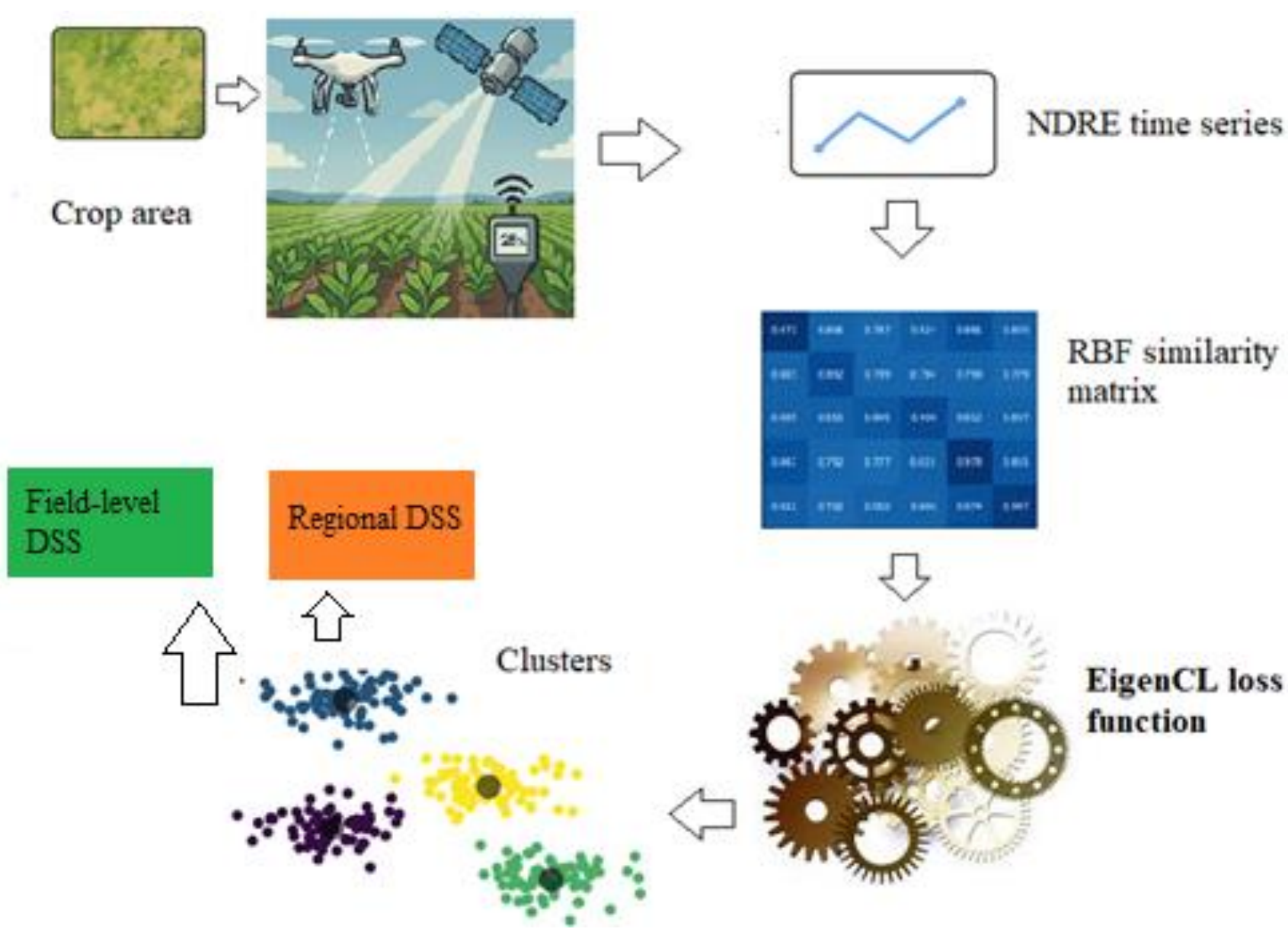

**Figure 1:** Workflow of EigenCL. Sentinel-2 NDRE time series are processed into a similarity matrix and optimized by EigenCL loss to form four stress clusters (Healthy, Mild, Moderate, Severe). These clusters can then be used for field-level decision support (e.g., scouting, irrigation, nutrient planning) or regional-scale monitoring (e.g., extension dashboards, crop insurance, policy evaluation)

## 2. Materials and Methods

**Table 1:** Summary of datasets used in this study. Iowa 2020 was used as the main training set. Nebraska 2023 tested cross-region transferability. Iowa 2020, 2022, and 2023 subsets were used for validation against soil moisture data.

| Data sets | Size (patches) | Role | Purpose |
|---|---|---|---|
| Iowa 2020 | 10000 | Training | Train EigenCL and assess internal cluster separability. |
| Nebraska 2023 | 3500 | Validation | Test cross-region generalization without retraining. |
| Iowa 2020 | 1500 | Validation | Compare clusters with soil moisture (Mesonet). |
| Iowa 2022 | 1500 | Validation | Compare clusters with soil moisture (Mesonet). |
| Iowa 2023 | 1500 | Validation | Compare clusters with soil moisture (Mesonet). |

### 2.1 Study Area and crop focus

The study focused on maize fields in Iowa and Nebraska, USA. These states represent major production zones that frequently experience drought stress. Iowa 2020 was selected as the primary training year because of widespread drought conditions. Nebraska 2023 was used as an external test case to evaluate transferability across regions.

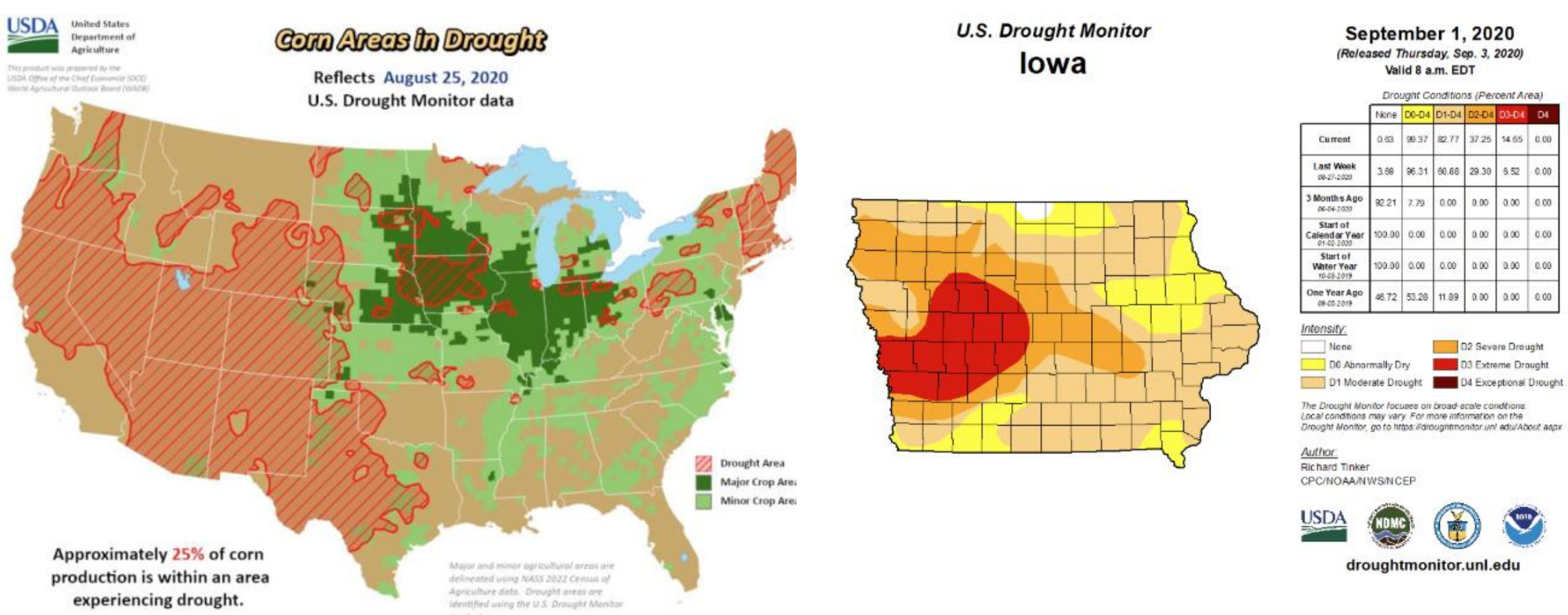

**Figure 2**: (a) US map highlighting drought-affected regions (b) Iowa drought region in 2020

### 2.2 Data sources (imagery, yield, soil, drought maps)

We used Sentinel-2 multispectral imagery (10 m resolution) to extract NDRE values. Patches of 100 × 100 pixels were sampled at five-day intervals across the growing season (July–September). A total of 10,000 patches were collected for Iowa 2020. For Nebraska 2023, 3,500 patches were sampled following the same protocol.

County-level yield data for 2019–2024 were obtained from USDA Quick Stats. These data were used to confirm that the selected stress years (Buena Vista 2020, Monona 2022, and Buchanan 2023) corresponded to lower yields compared with surrounding seasons.

Soil moisture data were retrieved from the Iowa Environmental Mesonet stations located near the sampled zones. We selected three stations: Masonville–Timeless Prairie Orchard in Buchanan County (TPOI4), Castana–Western ISU-RDF in Monona County (CNAI4), and Newell–Allee ISU-RDF in Buena Vista County (NWLI4). Daily soil moisture at 12–24-inch depth was used.

Drought-stressed zones were defined using the U.S. Drought Monitor maps for 2020, 2022, and 2023. These polygons ensured that validation was carried out in objectively stressed areas rather than arbitrary fields.

### 2.3 NDRE and trajectory extraction

For each patch, NDRE was calculated as:

$$NDRE = \frac{NIR - RE}{NIR + RE} \tag{1}$$

where NIR and RedEdge are the Sentinel-2 spectral bands. NDRE time-series were extracted at five-day cadence to capture crop trajectories across the season. These trajectories provide temporal signatures of chlorophyll dynamics rather than static values.

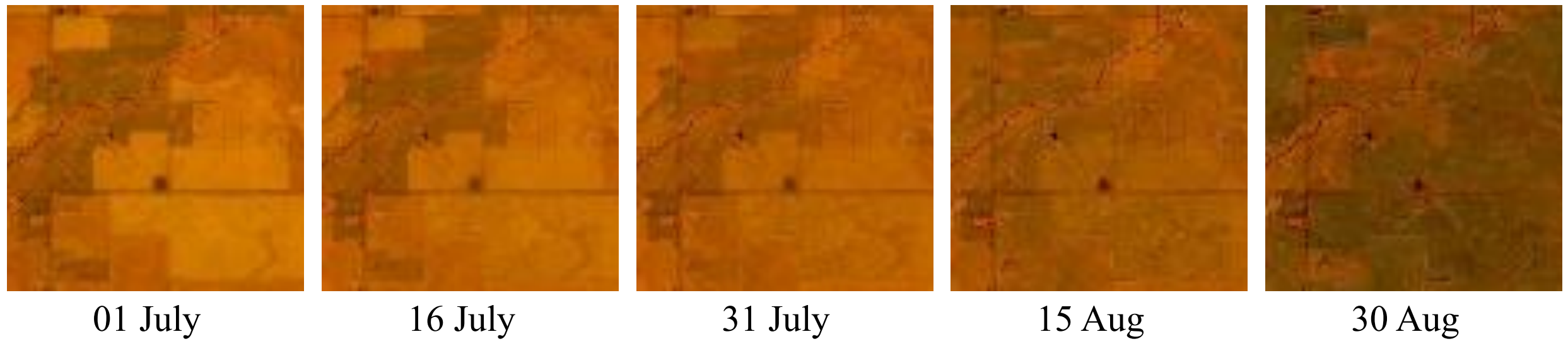

**Figure 3.** NDRE raster images of a single 100×100-pixel crop patch across five acquisition dates. This example illustrates how temporal vegetation index signals capture the progression of drought-induced crop stress. While Sentinel-2 NDRE is used here, EigenCL is not tied to a specific sensor or index and can generalize to other physiological signals.

### 2.4 EigenCL model design

To quantify temporal similarity between crop image patches based on their NDRE evolution, we constructed a Radial Basis Function (RBF) similarity matrix using the raw NDRE time series. Each image patch was represented as a five-dimensional vector $x_i \in R^5$, encoding mean NDRE values across five acquisition dates. Pairwise similarity between two image patches $x_i$ and $x_j$ computed using the Gaussian RBF kernel.

$$S_{ij} = \exp\left(-\gamma |x_i - x_j|^2\right) \tag{2}$$

Where $||x_i - x_j||^2$ denotes the squared Euclidean distance between the two NDRE time series, and $\gamma = \frac{1}{2\sigma^2}$ is the bandwidth parameter.

Each image patch was assigned a scalar eigen weight from this principal eigenvector, serving as a continuous and biologically meaningful proxy for stress severity. These eigen weights were later used to drive the push-pull mechanism in contrastive learning, enabling the emergence of stress-aware clusters within the learned embedding space.

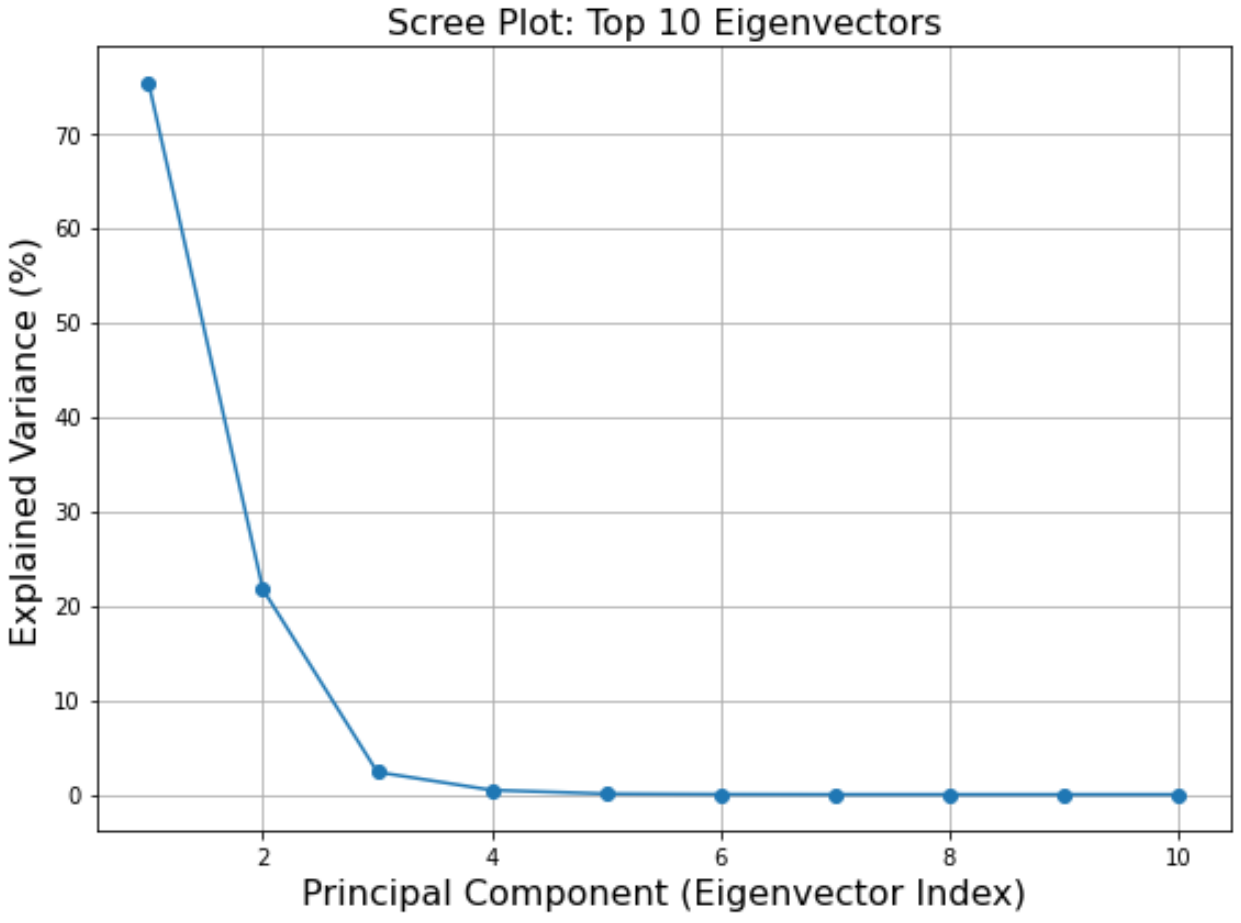

**Figure 4.** The plot shows how much variance is explained by each eigenvector. The first eigenvector alone captures a dominant share (~76%) of the total variance, justifying its use as the principal signal for guiding contrastive learning.

NDRE raster patches were input to a ResNet50 encoder to extract 2048-dimensional embeddings. These were projected through a nonlinear head and normalized. Instead of cosine similarity, embedding proximity was guided by stress similarity derived from eigenvector weights. This approach ensures that visually similar patches with differing stress responses are embedded distinctly, while patches with biologically similar stress profiles regardless of appearance are placed closer in the embedding space. These stress-aware embeddings form the foundation of the contrastive learning framework described in the following section.

To build an embedding space aligned with crop stress progression, we propose **EigenCL**, a physiological aware contrastive learning framework that replaces visual augmentation and cosine-based similarity with biologically grounded similarity derived from NDRE temporal patterns.

Each 2048-dimensional embedding obtained from ResNet50 is projected through a lightweight head comprising a fully connected layer with batch normalization and LeakyReLU activation. The projected embeddings $z_i \in R^d$ are L2-normalized and used to compute a cosine similarity matrix:

$$\mathrm{sim}\left(z_i, z_j\right) = \frac{z_i^\top z_j}{|z_i| \cdot |z_j|} \tag{3}$$

To guide the learning process, we extract the principal eigenvector $wi \in R$, from an RBF similarity matrix of the NDRE timeseries. This vector captures the dominant stress trajectory of each patch and is min-max normalized within each batch for numerical stability:

$$w_i = \frac{\max(w) - w_i}{\max(w) - \min(w)} \tag{4}$$

We define a physiologically informed similarity score $S_{ij}$ using an exponential decay function based on eigenvector weight differences, where similar stress trajectories yield higher similarity scores

$$\Delta_{ij} = \left|\widehat{w_i} - \widehat{w_j}\right|, \quad S_{ij} = \exp\left(-\frac{\Delta_{ij}}{\sigma}\right) \tag{5}$$

Here, $\sigma = 0.1$ is a smoothing hyperparameter. This formulation sharply penalizes dissimilar NDRE trajectories and constrains similarity scores to the range $[0,1]$.

The total loss consists of two components:

**a) Pull Loss (Biological Similarity-Based Attraction)**

For pairs with high similarity $S_{ij}$, the goal is to minimize their angular distance in the embedding space. We define the pull loss using a temperature-scaled logarithmic formulation:

$$L_{\text{pull}} = \sum_{i \neq j} S_{ij} \cdot \log\left(1 + \frac{1 - \text{sim}(z_i, z_j)}{\tau}\right) \tag{6}$$

where τ=0.075 is the temperature parameter, and the summation excludes self-pairs.

**b) Push Loss (Biological Dissimilarity-Based Repulsion)**

For dissimilar pairs (low $S_{ij}$), we apply a repulsion force using a margin-based hinge formulation. A weighting factor λ=4.0 amplifies the effect of dissimilarity:

$$L_{\text{push}} = \sum_{i \neq j} \lambda(1 - S_{ij}) \cdot \max(0, \text{sim}(z_i, z_j) - m) \tag{7}$$

where $m$=0.2 is the margin, encouraging dissimilar embeddings to remain at least margin (m) apart.

**c) Final Contrastive Loss**

The total loss is normalized by the number of unique pairs:

$$L_{\text{EigenCL}} = \frac{1}{N(N-1)} (L_{\text{pull}} + L_{\text{push}}) \tag{8}$$

where $N$ is the batch size (256 in our implementation).

Hyperparameters (λ, τ, σ, m) were tuned via grid search to maximize the Pearson correlation between embedding distances and NDRE-based stress trajectories on a held-out validation set.

This biologically guided contrastive formulation ensures that embeddings reflect stress progression rather than visual similarity alone. Visually similar patches with divergent NDRE profiles are separated in the embedding space, while biologically aligned patches regardless of appearance are clustered together, enabling detection of crop stress patterns.

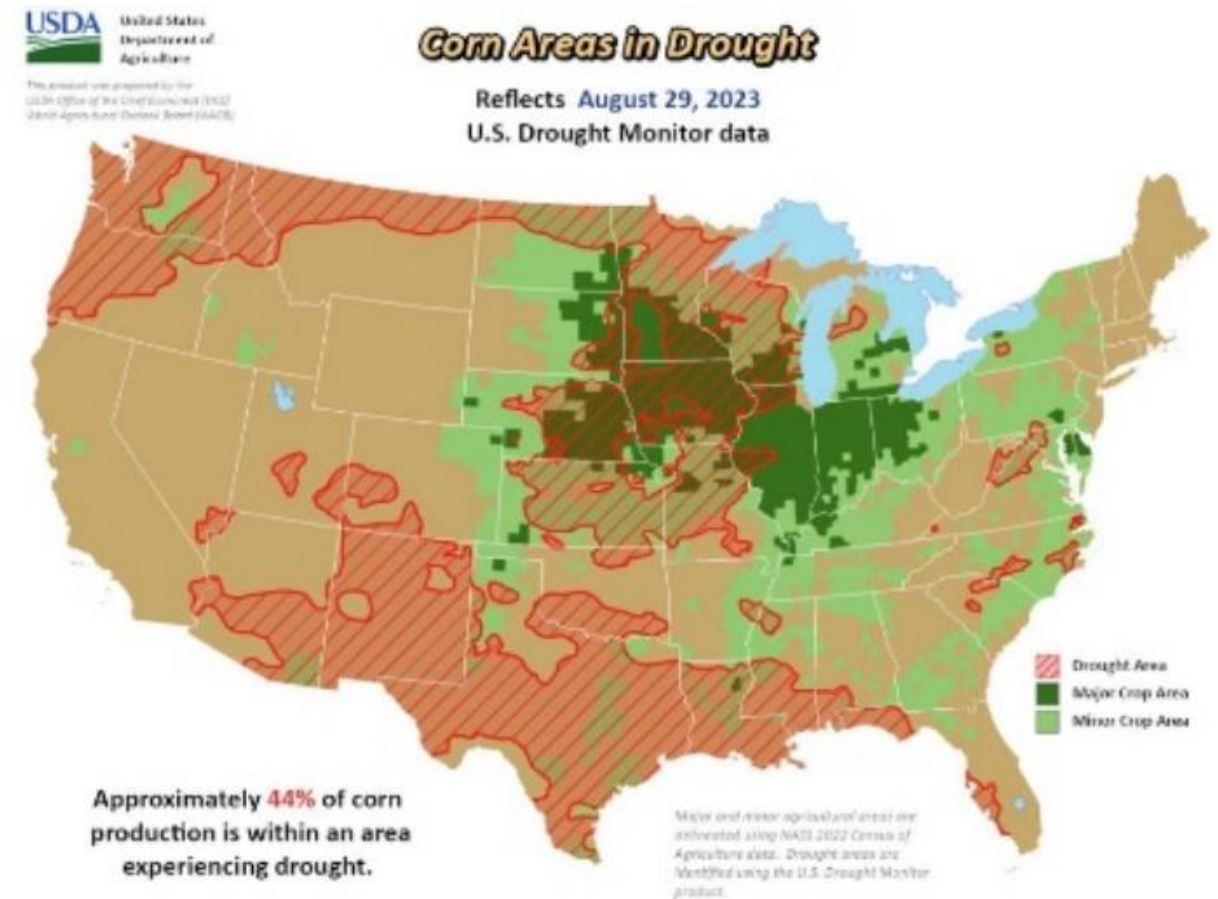

**Figure 5:** Drought region in Nebraska 2023.

To assess EigenCL's generalizability beyond the Iowa 2020 dataset, we conducted a secondary evaluation using Sentinel-2 NDRE imagery from the same cornfield in Nebraska, USA, during the 2023 growing season. This region experienced milder drought and a later onset of thermal stress. A total of 3,500 NDRE image patches were collected using the same five-date sampling protocol. The pre-trained model without any fine-tuning was directly applied using embeddings and cluster thresholds from the Iowa dataset.

We performed a grid search over key hyperparameters ($\lambda$, $\tau$, $\sigma$, m) using internal clustering metrics including Silhouette Score, Davies–Bouldin Index, and Calinski–Harabasz Index. Additionally, contrastive loss convergence and the stability of NDRE-based cluster centroids were considered to ensure physiological interpretability. The final configuration ($\lambda = 4.0$, $\tau = 0.075$, $\sigma = 0.5$, $m = 0.2$) consistently produced well-separated and biologically coherent stress clusters in both training and validation sets. Detailed results are provided in Appendix A.

## 3. Results and Discussion

To evaluate the quality and applicability of the proposed EigenCL framework, we conducted a comprehensive comparison against three alternatives: (i) an ablation model using cosine similarity without eigenvector guidance, (ii) K-Means applied to NDRE time series, and (iii) SimCLR, a generic contrastive learning method (Chen et al., 2020). All models shared the same ResNet-50 backbone to ensure consistent representation learning.

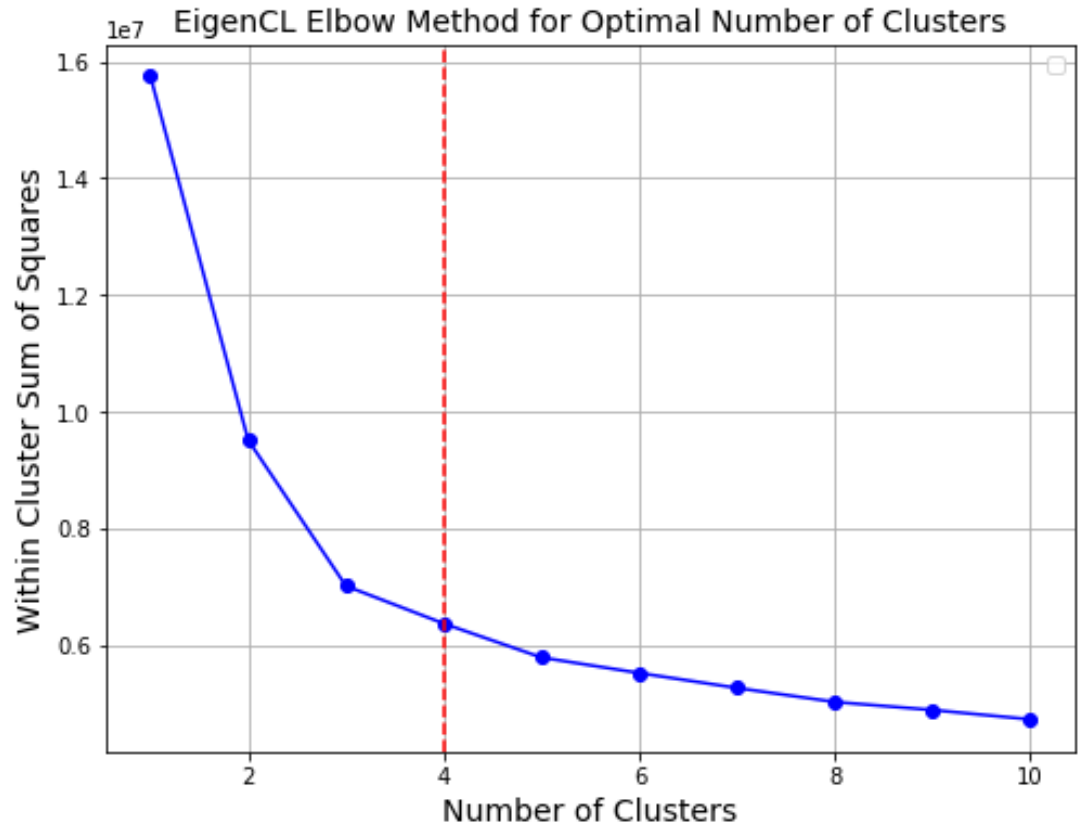

**Figure 6**: Elbow method to determine the optimal number of clusters from EigenCL embeddings.

### 3.1 Physiological and Agronomic Interpretation of Clusters

Clustering performance was assessed using Silhouette Score (higher indicates better separation), Davies–Bouldin Index (lower is better), and Calinski–Harabasz Index (higher is better). As summarized in Table 1, EigenCL outperformed all baselines with a Silhouette Score of 0.748, DBI of 0.350, and CHI of 49,624.06. The ablation variant exhibited moderate performance (Silhouette Score: 0.533), confirming the contribution of eigenvector guidance. SimCLR scored the lowest across all metrics, demonstrating limitations of generic contrastive learning for high-dimensional spectral-temporal data.

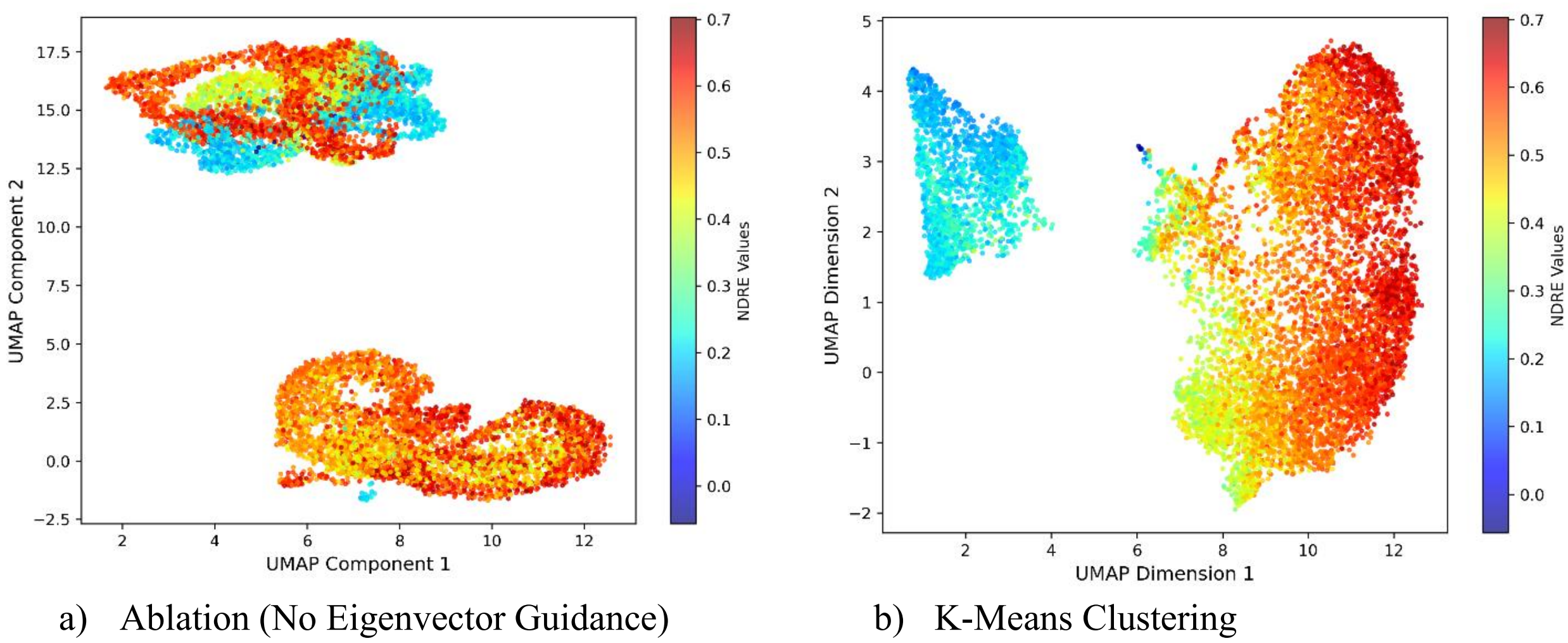

a) Ablation (No Eigenvector Guidance) b) K-Means Clustering

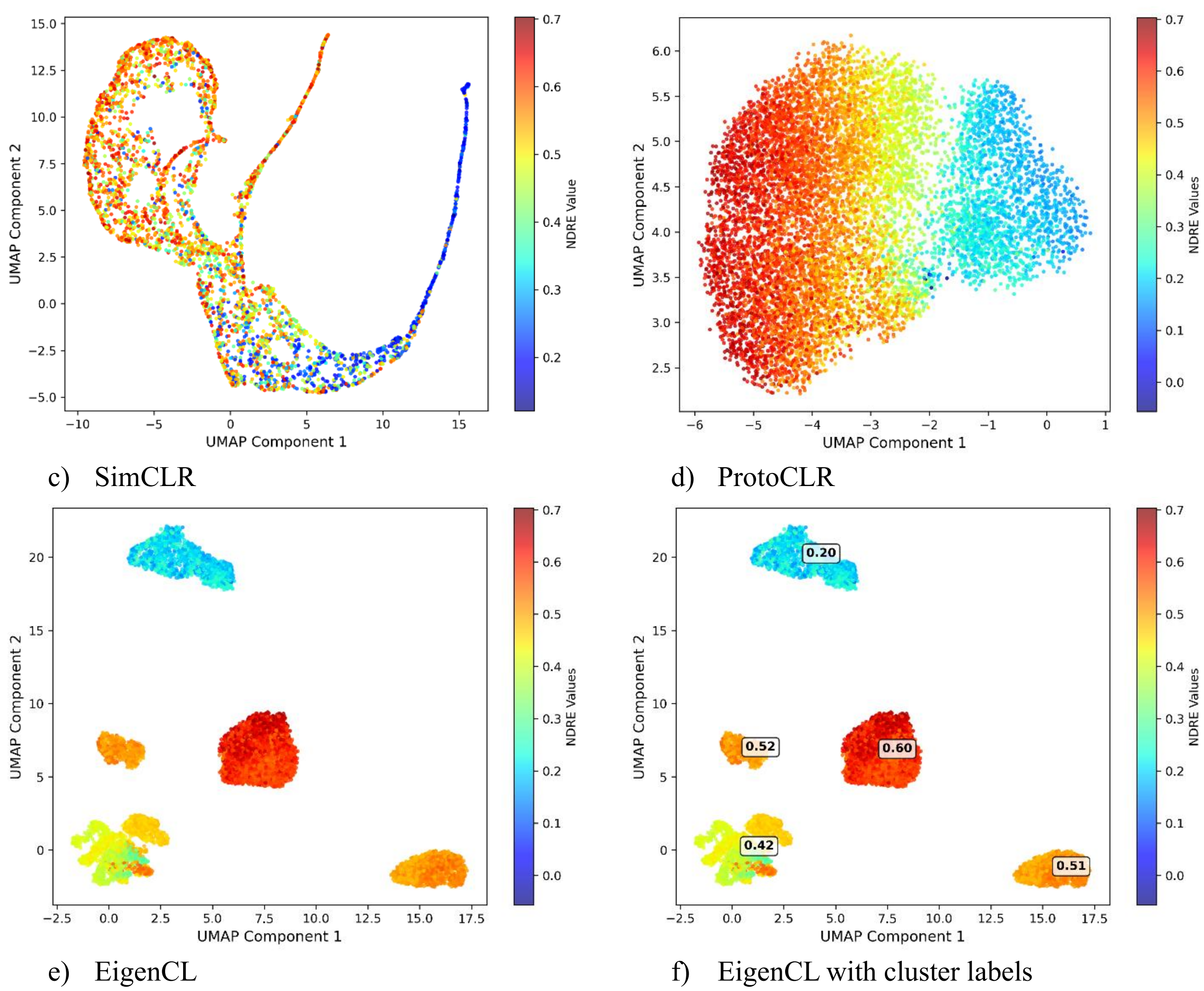

**Figure 7: Embedding and stress clustering comparison on the Iowa dataset.**
(a) UMAP projection of embeddings learned without stress-guided supervision. (b) NDRE time series clustered using classical K-Means (unsupervised baseline). (c) Clustering of embeddings learned via generic contrastive learning (SimCLR). (d) Prototype-based clustering using ProtoCLR with batch-defined prototypes. (e) Stress-aware clustering using eigenvector-guided contrastive learning (EigenCL), showing emergent structure driven by NDRE dynamics. (f) EigenCL clusters annotated with NDRE cluster mean values, illustrating physiological separability of stress stages.

**Table 2.** Clustering quality comparison across all models on Iowa dataset.

| Model | Backbone | Silhouette Score | Davies–Bouldin Index | Calinski–Harabasz Index |
|---|---|---|---|---|
| Ablation (Cos) | ResNet50 | 0.533 | 0.618 | 33445.60 |
| K-Means | ResNet50 | 0.465 | 0.958 | 19305.28 |
| SimCLR | ResNet50 | 0.416 | 0.724 | 2251.86 |
| ProtoCLR | ResNet50 | 0.348 | 0.892 | 9995.203 |
| **EigenCL** | **ResNet50** | **0.748** | **0.35** | **49624.06** |

To evaluate biological interpretability, UMAP projections of embeddings were overlaid with NDRE values, a well-established indicator of vegetation stress (Gitelson et al., 2005). NDRE was preferred over eigenvector weights due to its direct biophysical relevance and strong correlation (r = 0.95) with the guiding eigenvector. EigenCL produced well-separated clusters with internally consistent NDRE values, forming a clear monotonic gradient from low to high stress. Interestingly, clusters with similar mean NDRE values (e.g., 0.51 and 0.52) appeared distinct in UMAP space, reflecting divergent temporal patterns. This highlights EigenCL's ability to capture trajectory shape rather than just endpoint magnitude.

In contrast, the ablation model exhibited fractured clusters with moderate NDRE coherence, while K-Means produced statistically compact but biologically inconsistent groupings. NDRE values within K-Means clusters were heterogeneous, and no clear physiological gradient was observed. SimCLR, despite its popularity in computer vision, failed to form meaningful stress clusters, as confirmed by its low clustering metrics (Silhouette Score: 0.416, DBI: 0.724, CHI: 2,251.86) and high entropy in NDRE overlays. This underscores the necessity of domain-aware supervision in remote sensing time-series applications.

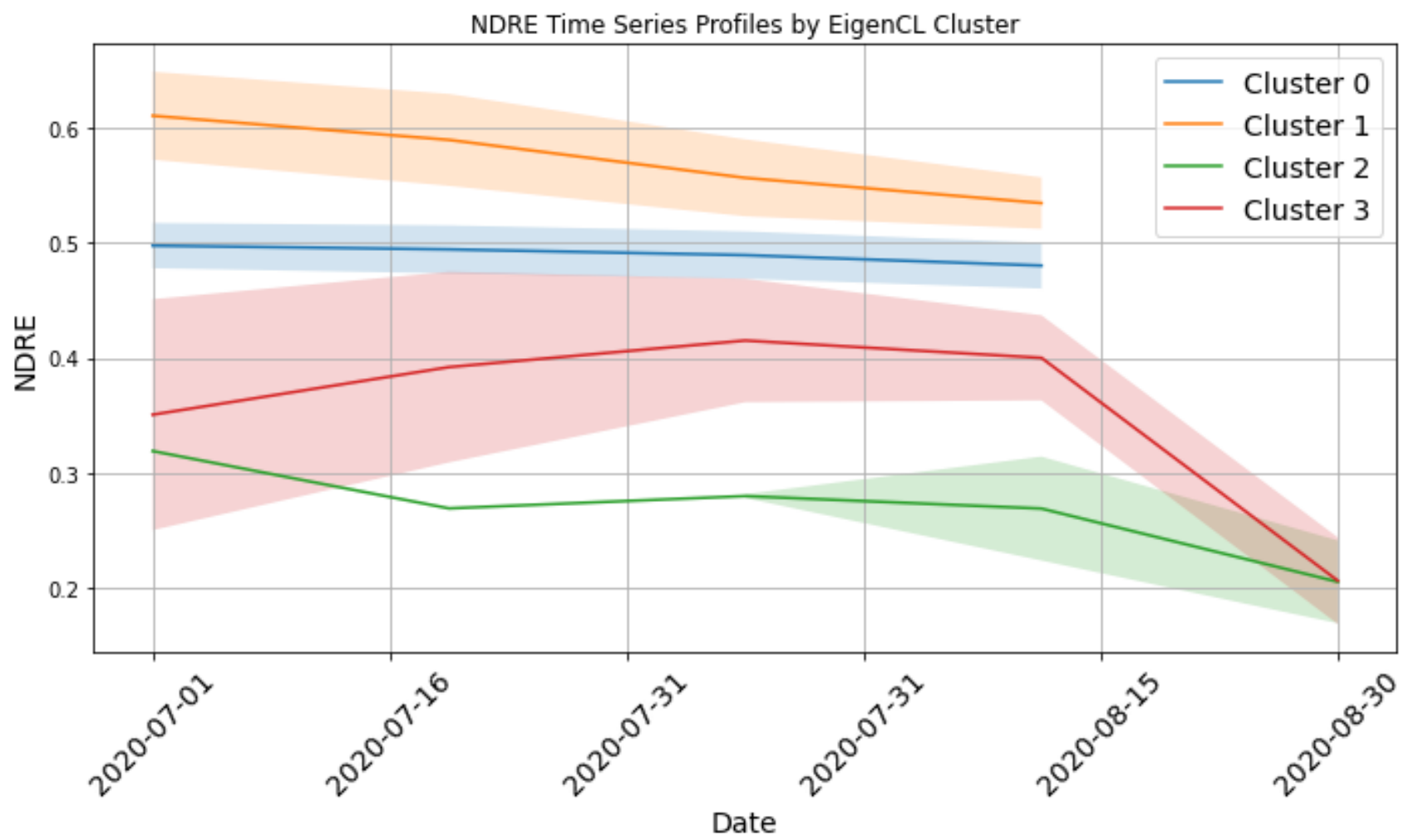

**Figure 8:** Mean NDRE time series profiles with 95% confidence intervals for each EigenCL-derived cluster. These distinct trajectories confirm that EigenCL captures physiological stress dynamics rather than superficial image similarity. While NDRE is used here as the signal, the framework can be extended to other physiological indicators.

NDRE profiles over time (Figure 8) show that EigenCL captures meaningful stress dynamics. Clusters differ clearly: Cluster 1 remains high (healthy), while Cluster 3 declines sharply (severe stress). This visual separation confirms that EigenCL clusters are not driven by static NDRE magnitude but by the trajectory shape, aligning with underlying stress dynamics. Such temporal coherence is critical for identifying stress detection and supports the model's biological interpretability.

To place the EigenCL-derived stress clusters in an agronomic context, we aligned the NDRE trajectories with typical maize growth calendars for Iowa and Nebraska (USDA-ERS, 2016; NC State Extension, 2024). Maize in these regions generally reaches rapid vegetative growth (V6–V8) in late June, tasseling and silking (VT–R1) around mid-July, and grain fill (R3–R5) in August. The EigenCL "severe stress" cluster showed sharp NDRE declines beginning in mid-July, coinciding with tasseling–silking, a stage widely recognized as the most drought-sensitive for maize yield formation (Daryanto et al., 2016). Clusters with moderate declines aligned with stress onset during late vegetative stages, while more gradual declines extended into grain fill, consistent with reduced kernel weight under late-season stress. These correspondences indicate that EigenCL is not only separating spectral trajectories but also capturing physiologically meaningful stress windows that directly influence yield outcomes.

### 3.2 Sensitivity to Eigenvector Selection

Replacing the principal eigenvector with lower-variance components during training led to a clear drop in clustering and classification performance. This confirms that the principal eigenvector capturing 76.3% of variance and strongly correlated with NDRE is essential for guiding biologically meaningful embeddings. We therefore retain it as the sole supervisory signal in EigenCL.

### 3.3 Downstream Utility

To assess the utility of EigenCL embeddings, we trained two classifiers k-Nearest Neighbors and Logistic Regression on the frozen embeddings without fine-tuning. On a 70/30 split, k-NN achieved 89.1% accuracy (F1: 0.87), and Logistic Regression reached 85.2% (F1: 0.82). These results confirm that the embeddings encode both linear and nonlinear separable patterns relevant for stress classification.

**Table 3**: Downstream Classification Performance Using EigenCL Embeddings

| Classifier | Accuracy (%) | F1 (Macro) | Precision (Macro) | Recall (Macro) |
|---|---:|---:|---:|---:|
| k-Nearest Neighbors | 89.1 | 0.87 | 0.86 | 0.88 |
| Logistic Regression | 85.2 | 0.82 | 0.81 | 0.84 |

### 3.4 Statistical Confirmation of Cluster Separation

A one-way ANOVA confirmed that NDRE means varied significantly across EigenCL clusters ($F = 37,950.39$, $p < 0.0001$). Tukey's HSD test revealed all pairwise differences were statistically significant ($p < 0.0001$), validating that each cluster represents a distinct physiological regime. The largest difference (–0.398) was between Clusters 0 and 2, while even the smallest (–0.076) between Clusters 1 and 3 remained significant.

**Table 4**: Statistical Comparison of NDRE Means Across Detected Clusters

| Cluster Pair | Mean NDRE Difference | p-value |
|---|---:|---:|
| Cluster 0 vs 2 | –0.398 | < 0.0001 |
| Cluster 1 vs 2 | –0.314 | < 0.0001 |
| Cluster 2 vs 3 | 0.237 | < 0.0001 |

For the preliminary study the pretrained model without any fine-tuning was directly applied on Nebraska datasets. Despite environmental differences, EigenCL maintained consistent performance, retaining stress-aware clustering and NDRE interpretability. While a modest decline in clustering metrics was observed, this was expected due to regional variability. Nonetheless, the trends remained robust, validating EigenCL's ability to generalize across time and geography. These results support its deployment in longitudinal monitoring and scalable precision agriculture workflows.

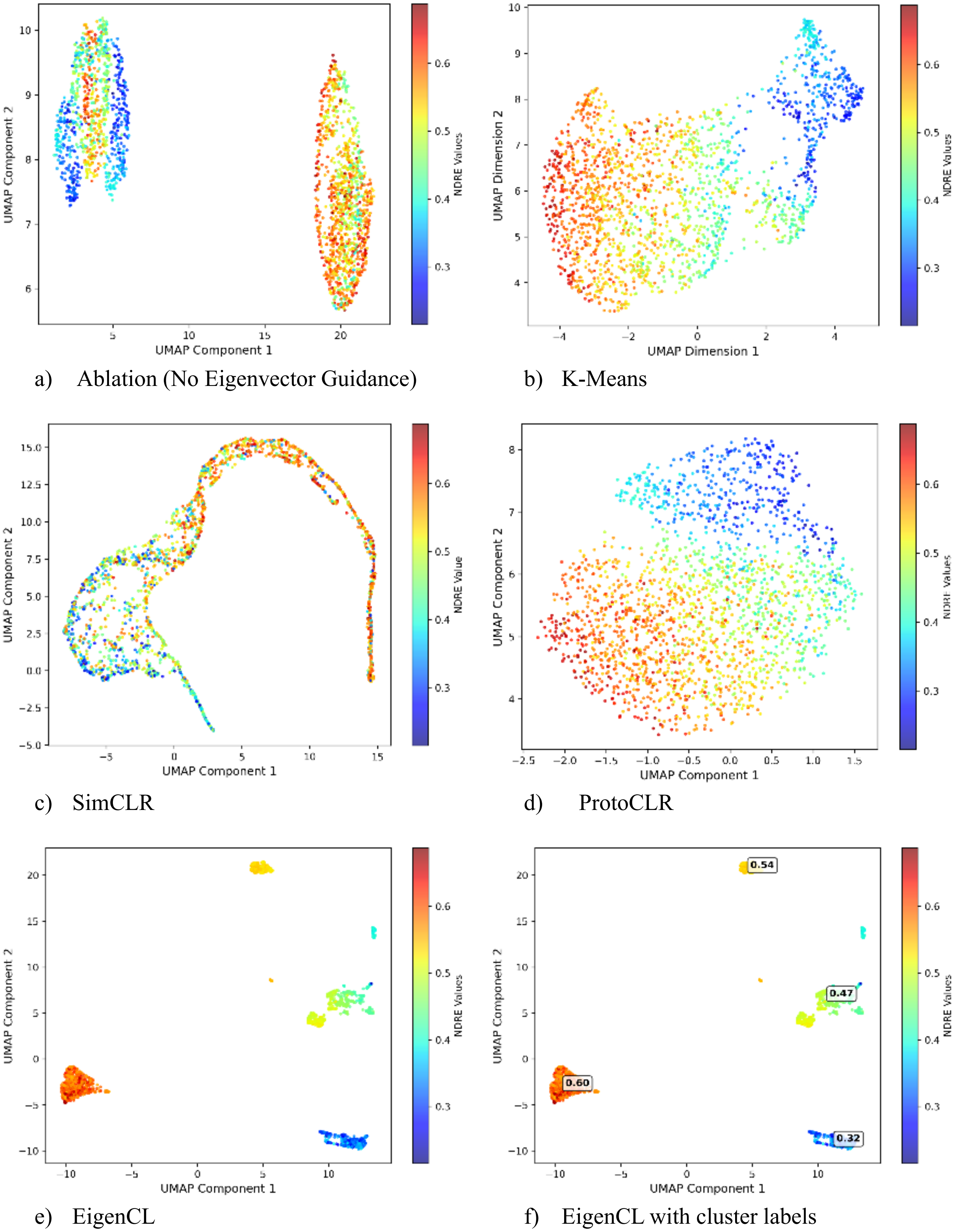
UMAP Component 1
UMAP Component 2
NDRE Values
a) Ablation (No Eigenvector Guidance)
UMAP Dimension 1
UMAP Dimension 2
NDRE Values
b) K-Means
UMAP Component 1
UMAP Component 2
NDRE Value
c) SimCLR
UMAP Component 1
UMAP Component 2
NDRE Values
d) ProtoCLR
UMAP Component 1
UMAP Component 2
NDRE Values
e) EigenCL
0.54
0.47
0.60
0.32
UMAP Component 1
UMAP Component 2
NDRE Values
f) EigenCL with cluster labels

**Figure 9:** Embedding and stress clustering comparison on the Nebraska dataset.
(a) UMAP projection of embeddings learned without stress-guided supervision. (b) NDRE time series clustered using classical K-Means (unsupervised baseline). (c) Clustering of embeddings learned via generic contrastive learning (SimCLR). (d) Prototype-based clustering using ProtoCLR with batch-defined prototypes. (e) Stress-aware clustering using eigenvector-guided contrastive learning (EigenCL), showing emergent structure driven by NDRE dynamics. (f) EigenCL clusters annotated with NDRE cluster mean values, illustrating physiological separability of stress stages.

**Table 5**: County corn yields (bu/acre; 2019–2024; USDA Quick Stats) for Buena Vista, Monona, and Buchanan counties. The selected stress years used in this study Buena Vista (2020), Monona (2022), and Buchanan (2023) correspond to the lowest yields within their respective counties compared with surrounding seasons, supporting their designation as drought-affected years.

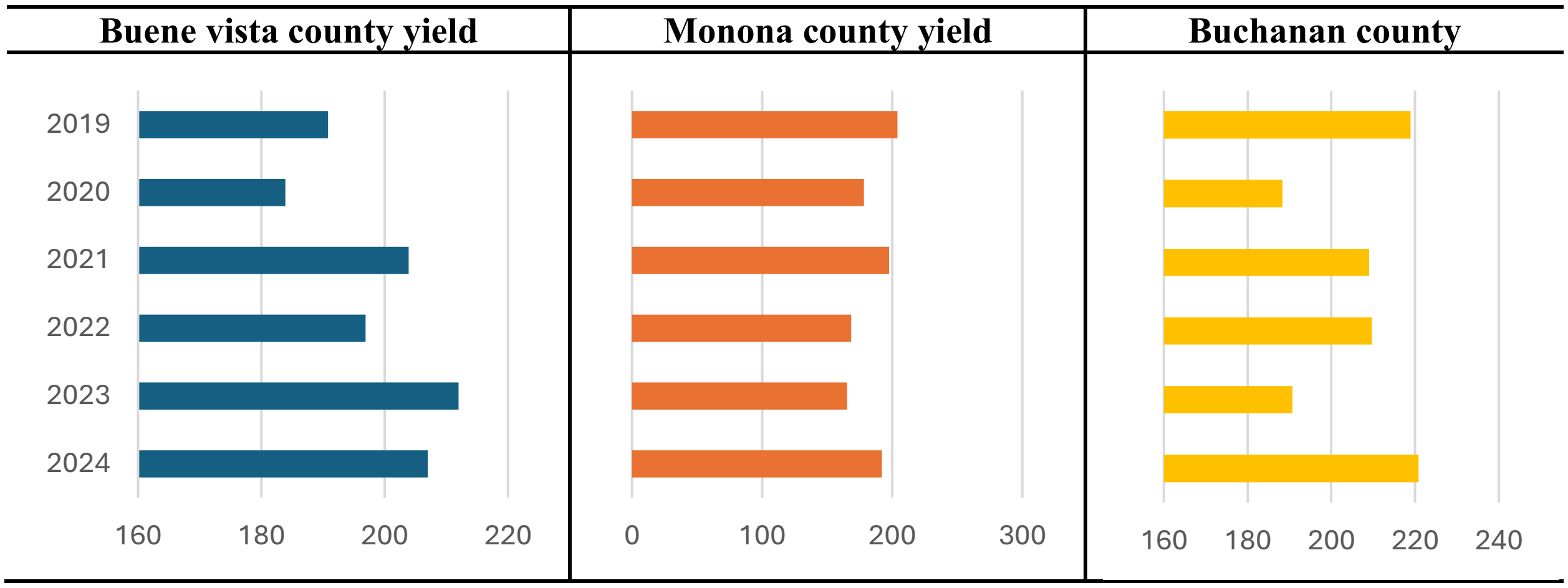

### 3.5 Validation with soil moisture, drought maps, and yields

We built a ground-truth dataset for three Iowa seasons by using the U.S. Drought Monitor maps to locate drought-stressed zones for 2020, 2022, and 2023. To confirm that these years were indeed stress years, we also cross-checked county-level yield data (Table 5; USDA Quick Stats), which shows that Buena Vista (2020), Monona (2022), and Buchanan (2023) recorded noticeably lower yields compared with surrounding seasons. We then anchored sampling to three Iowa Environmental Mesonet stations situated in those zones: Masonville–Timeless Prairie Orchard in Buchanan County (TPOI4), Castana–Western ISU-RDF in Monona County (CNAI4), and Newell–Allee ISU-RDF in Buena Vista County (NWLI4). Around each station we defined a ~50 km² window and extracted Sentinel-2 NDRE on the sensor's five-day cadence; for the same dates we pulled daily soil-moisture records from the Mesonet site. Each county window contains approximately 600–900 agricultural fields, providing diverse cropping and management conditions for evaluation. Using this paired record, we formed soil-based stress classes (per-year quartiles, lower moisture = higher stress) as ground truth and, in parallel, ran EigenCL on the NDRE series with unchanged parameters to obtain four NDRE clusters (Table 4). We then compared NDRE clusters with soil-based classes, including lag analyses where soil leads NDRE by a few days, to quantify agreement.

**Table 6**: EigenCL driven NDRE Clusters by year (Iowa; 2020, 2022, 2023)

| Year | Healthy | Mild Stress | Moderate Stress | Sever Stress |
|---|---|---|---|---|
| 2020 | 0.650 | 0.520 | 0.4050 | 0.330 |
| 2021 | 0.470 | 0.390 | 0.3230 | 0.230 |
| 2023 | 0.410 | 0.350 | 0.2950 | 0.180 |

### 3.5 Lagged NDRE–soil moisture correlation

Because canopy reflectance responds after changes in soil water, we tested lagged agreement by correlating NDRE at day t with 12-inch soil moisture measured d days earlier, for d = 0 to 14 days. Observations were paired by their recorded dates at each lag. For every lag we computed Spearman rank correlation with bootstrap 95% confidence intervals and permutation p-values. The best lag is the d with the largest absolute Spearman correlation. For classification agreement, we shifted soil by the best lag and then derived soil labels (per-year quartiles; lower VWC means higher stress) before computing ARI between NDRE labels and soil labels. Same-day results (lag 0) are reported as a baseline.

Peak agreement occurred at 6 days in 2020 (Spearman rho = 0.525, 95% CI [0.319, 0.667], p = 0.002), 14 days in 2022 (rho = 0.721, 95% CI [0.519, 0.831], p = 0.002), and 0 days in 2023 (rho = 0.608, 95% CI [0.444, 0.734], p = 0.002). Pearson correlations showed the same pattern (2020: r = 0.514; 2022: r = 0.608; 2023: r = 0.524). Overall, NDRE tracks soil moisture within about zero to two weeks, with the exact lag varying by year.

**Table 7.** Best-lag correlation summary (soil leads NDRE by d days)

| Year | Best lag (days) | Spearman rho | 95% CI | p-value | Pearson r |
|---|---|---|---|---|---|
| 2020 | 6 | 0.525 | [0.319, 0.667] | 0.002 | 0.514 |
| 2022 | 14 | 0.721 | [0.519, 0.831] | 0.002 | 0.608 |
| 2023 | 0 | 0.608 | [0.444, 0.734] | 0.002 | 0.524 |

### 3.6 Classification agreement (ARI) between NDRE clusters and soil-derived

Soil-based stress classes were treated as ground truth. For each lag d, we first shifted soil by d days, then assigned soil labels using per-year quartiles of 12-inch VWC (lower moisture means higher stress). We computed ARI between the EigenCL NDRE labels and these soil labels for the same dates. We report the best lag and the same-day baseline.

Using the lag that best aligns soil and NDRE improves agreement versus same-day labels. In 2020 the best lag was 6 days with ARI = 0.120 (lag 0 = 0.013). In 2022 the best lag was 8 days with ARI = 0.155 (lag 0 = 0.014). In 2023 the best lag was 2 days with ARI = 0.371 (lag 0 = −0.006). The improvement is most pronounced in 2023.

**Table 8.** ARI summary (best lag vs same day)

| Year | Best lag for ARI (days) | ARI (best) | ARI (lag 0) |
|---|---|---|---|
| 2020 | 6 | 0.120 | 0.013 |
| 2022 | 8 | 0.155 | 0.014 |
| 2023 | 2 | 0.371 | −0.006 |

### 3.7 EigenCL as a Decision Support Tool

Operational use of EigenCL. EigenCL can be used in practical farm and regional decision systems. At the field scale, the four stress clusters can be shown as heatmaps (Healthy, Mild, Moderate, Severe), highlighting where stress is starting and where it is already severe. These maps help farmers decide where to scout, apply irrigation, or adjust fertilizer. At the regional scale, the share of area in each cluster can be summarized into a stress risk index for extension services, cooperatives, or crop insurers to monitor stress levels across counties. In this way, EigenCL converts raw NDRE time-series into simple, interpretable alerts and maps for decision-makers. A schematic workflow is shown in Figure

### 3.8 Using Clusters for DSS Rules

Because EigenCL outputs four clear stress levels, the DSS can link clusters to if–then rules:

- Severe (e.g., ≥ 20% of field patches): send an alert for urgent scouting and, where feasible, supplemental irrigation within 24–48 h.
- Moderate (or Severe+Moderate ≥40%): prioritize scouting, check soil moisture/valves, and consider variable-rate input plans.
- Mild (e.g., 20–40% with limited spread): monitor weekly; no immediate action, but flag for follow-up.
- Healthy (majority of area): no action; continue routine monitoring.

At the regional level, aggregate cluster shares into a simple index, for example a 0–3 weighted score (Healthy=0, Mild=1, Moderate=2, Severe=3) averaged over fields or pixels to track stress risk by county and week. These rules turn EigenCL clusters into operational signals at both farm and regional scales.

### 3.9 Linking Stress Clusters to Yield and Management Outcomes

The EigenCL clusters are not only patterns in NDRE, they relate to yield outcomes. Counties and seasons with higher proportions of Severe clusters showed larger negative yield anomalies than those dominated by Healthy clusters, consistent with known maize physiology, where stress around tasseling-silking (R1) strongly reduces yield. Early detection of fields shifting into Moderate/Severe clusters can support timely responses (e.g., irrigation during R1 where available) and inform insurance or regional risk decisions. While per-cluster yield penalties are not yet quantified, the cross-year and cross-region robustness suggests strong potential to link EigenCL outputs to yield forecasting and farm risk management tools in future work

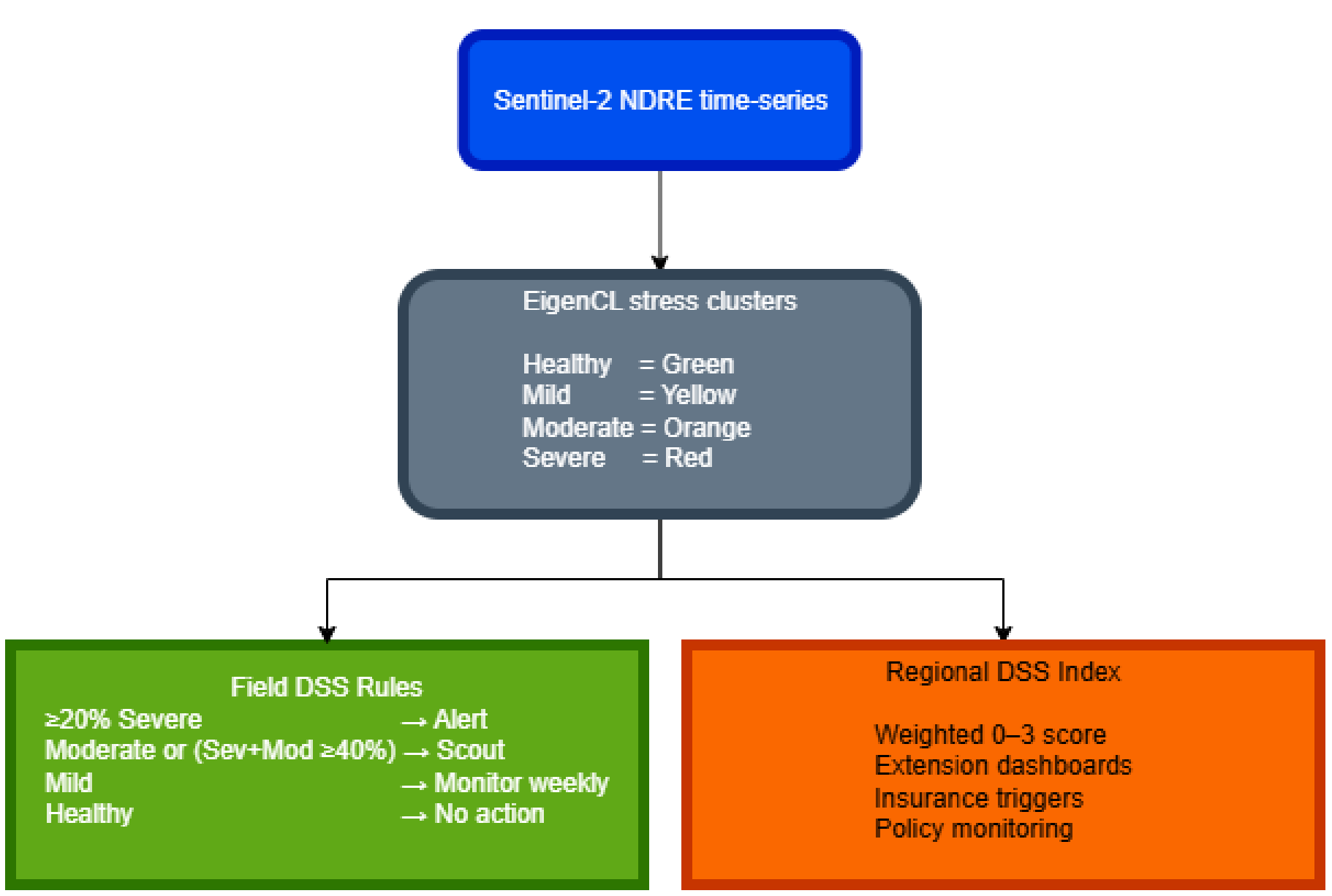

**Figure 10:** Workflow for integrating EigenCL into a Decision Support System (DSS). Sentinel-2 NDRE time-series are processed by the EigenCL model to produce four stress clusters (Healthy, Mild, Moderate, Severe). Outputs can then be used at the field level to trigger scouting or irrigation actions through cluster-based rules, and at the regional level to build stress indices for extension dashboards, crop insurance, and policy monitoring.

## 4. Limitations and Future Work

Although EigenCL demonstrated robustness across seasons and regions, the present study is limited to maize in Iowa and Nebraska and does not yet cover a wider range of crops, agroecological zones, or longer historical periods. In addition, while stress clusters were linked to soil moisture, drought maps, and county-level yield anomalies, precise quantification of yield penalties at the field scale remains an open step. Future work should extend EigenCL to integrate additional signals such as thermal indices, hyperspectral features, and in-field sensor data, and evaluate its performance on multi-crop, multi-region datasets. Finally, operational deployment will require embedding EigenCL outputs into practical DSS platforms, delivering cluster-based maps, alerts, and risk indices that can guide farm management, extension services, and crop insurance.

## 5. Conclusion

This study presented EigenCL, a physiology-aware learning framework that stages crop stress from NDRE trajectories as indicators of canopy chlorophyll dynamics. Tested across maize fields in Iowa (2020) and Nebraska (2023), EigenCL produced four stress clusters that aligned with soil-moisture patterns, drought monitor maps, and yield anomalies, demonstrating both physiological coherence and agronomic relevance. By embedding physiological processes into data-driven modeling, EigenCL offers an interpretable and transferable approach to crop stress diagnostics. The framework supports field-scale decision-making through stress maps and alerts that can guide irrigation or nutrient management, and provides regional stress indices for extension, insurance, and policy monitoring. Although demonstrated here with Sentinel-2

NDRE, EigenCL is sensor-agnostic and adaptable to UAV and proximal data streams, making it scalable across farms and landscapes. Expanding validation to other crops, agroecological zones, and multi-sensor datasets will further consolidate its role as a foundation for decision support systems that link remote sensing to sustainable and climate-smart agronomy.

**Data Availability**

The data that support the findings of this study are openly available: Iowa NDRE patches (Kaggle: https://www.kaggle.com/datasets/shafqaatahmad/crop-satellite-images-with-ndre-spectral-band) and Nebraska 2023 NDRE patches (Kaggle: https://www.kaggle.com/datasets/shafqaatahmad/ndre-crop-stress-dataset-nebraska-2023-corn).Code for EigenCL training and inference is available from the corresponding author upon reasonable request.

**Disclosure Statement**

The authors declare that there are no conflicts of interest related to this study.

**Appendix A**. Hyperparameter Grid Search Summary

**Table A1**. Summary of hyperparameter tuning for the EigenCL framework.

| Parameter | Values Tested | Best Value | Criteria Used for Selection |
|---|---|---|---|
| $\lambda$ | 1.0, 2.0, 4.0, 6.0 | 4 | Silhouette Score, NDRE gradient stability, loss convergence |
| $\tau$ | 0.05, 0.075, 0.1, 0.15 | 0.075 | Silhouette Score, DBI, contrastive loss value |
| $\sigma$ | 0.3, 0.5, 0.7, 1.0 | 0.5 | CHI, smooth loss trajectory, cluster cohesion |
| m | 0.1, 0.2, 0.3 | 0.2 | Visual cluster separation, stable NDRE centroids |